\documentclass{llncs}

\usepackage[cmex10]{amsmath}
\usepackage{amsfonts}
\usepackage{algorithmic}
\usepackage[ruled,vlined,linesnumbered,nokwfunc]{algorithm2e}
\usepackage{array}

\usepackage{multirow}
\usepackage{textcomp}

\usepackage{color}

  \usepackage[pdftex]{graphicx}
   \graphicspath{{img/pdf/}{jpeg/}{img/}}
    \DeclareGraphicsExtensions{.pdf,.jpeg,.png}

\newcommand{\DB}{{\it DB}}
\newcommand{\A}{\mbox{\bf A}}

\newcommand\Dom{\mathbb{D}}

\newcommand\out[2]{{\rm out}^{#1}_{#2}}
\newcommand\Out{\out{}{a}}

\begin{document}

\title{Outlying Property Detection \\ with Numerical Attributes}

\author{Fabrizio Angiulli\inst{1} \and Fabio Fassetti\inst{1} \and Giuseppe Manco\inst{2}
\and Luigi Palopoli\inst{1}}
\institute{
DIMES Dept., University of Calabria, Rende, Italy\\
\texttt{$\{$f.angiulli,f.fassetti,palopoli$\}$@dimes.unical.it}
\and
ICAR-CNR, Rende, Italy\\
\texttt{manco@icar.cnr.it}
}

\date{}

\maketitle

\begin{abstract}
The {\em outlying property detection problem} is the problem of
discovering the properties distinguishing 
a given object, known in advance to be an outlier in a database, from the other
database objects.
In this paper, we analyze the problem
within a context where numerical attributes are taken into account,
which represents a relevant case left open in the literature.
We introduce a measure to quantify the degree the outlierness of an
object, which is associated with the relative likelihood of the value, 
compared to the to the relative likelihood of other objects in the
database. 
As a major contribution, we present an efficient algorithm to compute
the outlierness relative to significant subsets of the data. The
latter subsets are characterized in a “rule-based” fashion, and hence
the basis for the underlying explanation of the outlierness. 
\end{abstract}

\section{Introduction}

In this work we aim at \emph{characterizing} outliers. 
\emph{Outliers} are the \textit{exceptional} objects
in the input dataset, that is to say 
objects that significantly differ from the
rest of the data. 
Approaches to outlier detection introduced in the literature
can be classified in {supervised} \cite{ChawlaJK04},
which exploit a training set of normal and abnormal objects,
{semi-supervised} \cite{ScholkopfBV95}, which assume that only normal examples are given,
and {unsupervised} \cite{BL94,KN98,BKNS00,PKGF03,AP05,AP06,LiuTZ08,AngiulliF09,ChandolaBK09}, 
which search for outliers in an unlabelled data set.

It is worth to notice that 
the above mentioned methods focus only on \emph{identification}, and they do not
concentrate on providing a \emph{description} or an \emph{explanation} of why
an identified outlier is exceptional, which is vice versa the problem we 
are intended to face here.
While outlier detection in datasets has been one of the most
widely investigated problems in data mining, the related problem of
outlier explanation received less attention in the literature.
It must be noticed that the outlier explanation problem
is completely different
from supervised and semi-supervised outlier detection, and, moreover,
is to be considered
orthogonal to the unsupervised outlier detection task.

As an example of outlier explanation task, assume you are analyzing health parameters of a sick
patient, which include several features such as body
temperature, blood pressure measurements and others. If an history of
healthy patients is available,
then it is relevant to single out those parameters that
mostly differentiate the sick patient from the healthy population.
It is important to highlight here that the abnormal individual,
whose peculiar characteristics we want to detect, is provided
as an input to the outlier explanation problem, that is, this individual has been recognized
as anomalous in advance by the virtue of some external information,
mean or procedure.

The focus of this paper is the discovery of \textit{outlying properties}:  we
are interested in unveiling the hidden structures that 
make an input outlier object $o$ special w.r.t. an input population. 
This can be accomplished
($i$) by detecting the subsets $S$ of the input population 
that represent an homogeneous sub-population, 
intuitively a set of objects sharing similar features
(which we will refer to as \emph{explanations}),
including $o$, and
($ii$) by identifying attributes 
(also referred to as \emph{properties})
where $o$ substantially differentiates from the other objects in $S$.

With this aim,
subspace outlier mining techniques, like the one presented in \cite{AY01}, 
could in principle be used to extract information about outlier
properties. However, the originary task considered in \cite{AY01}
is different from
the task investigated here, since subspaces therein
highlight the outlierness, whereas in our approach they represent an
homogeneous subpopulation upon which to compare a given property. 
In \cite{KN99}, the authors focus on the
identification of the intensional knowledge associated with
distance-based outliers. 
However, this setting models outliers which are exceptional
with respect to the whole population, but it does not capture objects
that are exceptional only  if compared to homogeneous
subpopulations. 
In \cite{AngiulliFP13} an outlier subpopulation is given in input
and compared with the inlier population in order to simultaneously 
characterize the whole exceptional subpopulation.
Despite the latter approach shares with ours a common rationale,
we note that the framework considered in \cite{AngiulliFP13}
is different and the solutions there proposed
cannot be applied to the special case in which the outlier sub-population
consists of just one single individual, which is precisely
the scenario considered here.

A viable solution to the outlying property
detection problem has been
devised in  \cite{AngiulliFP09}. 
Specifically, 
a set of attributes witnesses the abnormality of an object if the combination
of values the object assumes on these attributes is very infrequent
with respect to the overall distribution of the attribute values in
the dataset, and this is measured my means of the so called
\emph{outlierness} score.
A major problem with the outlierness score presented in
\cite{AngiulliFP09} is that it was specifically designed and shown
effective for categorical attributes. Hence the question is how to
adapt that idea to a more general setting with both
categorical and numerical attributes.
We point out that 
discretizing numerical attributes and applying the technique 
of \cite{AngiulliFP09} to
the discretized attributes is not a suitable solution, for several
reasons. 
First of all, the result of the analysis
will strongly depend on the results of the discretization process.
This drawback is further exacerbated by
the peculiarities of the outlierness measure, which
assigns higher scores to  very unbalanced distributions, and by
contrast provides low scores to uniform frequency distributions. 
In a sense, the discretization process should be supervised by the outlierness score,
in order to detect in the first place the bins capable of magnifying the score itself.

The appropriate treatment of numerical attributes is indeed one
of the main problems we deal with in this paper.
Specifically, the main contribution of this work amounts to provide an
\textit{outlierness} measure representing a refined generalization of
that proposed in \cite{AngiulliFP09} and which is able to quantify the
exceptionality of a given numerical or categorical \textit{property} featured by the given
input anomalous object with respect to a reference data population.
In particular, in order to quantify the degree of
unbalanceness between the frequency of the value under
consideration and the frequencies of the rest of the database values,
our measure analyzes the
curve of the cumulative distribution function (\textit{cdf})
associated with the occurrence probability of the domain values.  
It is worth noting that relying on the
\textit{cdf} allows to correctly recognize exceptional properties
independently of the form of the underlying probability density
function (\textit{pdf}), since the former compares the occurrence
probabilities of the domain values rather than directly comparing the
domain values themselves.
This enables us to build 
a general methodology for uniformly mining exceptional properties in the
presence of both categorical and numerical attributes, so that a fully
automated support is provided to decode those properties determining
the abnormality of the given object within the reference data context.

The rest of the paper is organized as follows.
Section \ref{sect:outlierness0} introduces the outlierness measure
and the concept of explanation.
Section \ref{sect:algo} describes the method for computing outlierness
and determining associated explanations.
Section \ref{sect:experim} discusses experimental results.
Finally, Section \ref{sect:concl} presents conclusions
and discusses future work.

\section{Outlierness and Explanations}\label{sect:outlierness0}

To begin with, we fix some notation to be used throughout the paper. 
In the following, $a$ denotes an {\em attribute},
that is an identifier with an associated domain $\Dom(a)$,
and ${\bf A}=a_1,\ldots,a_m$ denotes a set of $m$ attributes.
The {\em value} $v_i$
associated with the attribute $a_i$ in the object $o$ will be denoted by
$o[a_i]$. A {\em database} $\DB$ on a set of attributes $\bf A$ is a
multi-set of objects on $\bf
A$.

We shall characterize populations in a
``rule-based'' fashion, by denoting the subset of $\DB$ that embodies
them.
Formally, a \emph{condition} on $\bf A$ is an expression of the form $a \in [l,u]$, where
($i$) $a\in {\bf A}$, ($ii$) $l,u \in \Dom(a)$, and ($iii$) $l\le u$, if $a$ is
numeric, and $l=u$, if $a$ is categorical.
If $l=u$, the interval $I=[l,u]$ is sometimes abbreviated as $u$ and the
condition as $a\in I$ or $a=I$.
Let $c$ be a condition $a\in [l,u]$ on $\bf A$.
An object $o$ of $\DB$ satisfies the condition $c$,
if and only if $o[a]$ equals $l$, if $a$ is categorical, or 
$l \le o[a] \le u$, if $a$ is numerical.
Moreover,
$o$ satisfies
a set of conditions $C$ if and only if $o$ satisfies each condition $c\in C$.
Given a set $C$  of conditions on $\bf A$.
The \textit{selection} $\DB_C$ of the database $\DB$
w.r.t. $C$ is the database consisting of the objects $o\in\DB$
satisfying $C$.

Next, the definitions of outlierness 
and explanation 
are introduced.

\subsection{Outlierness}
\label{sect:outlierness}
This 
measure is used to quantify the exceptionality of a property. 
The intuition underlying this measure is that an
attribute makes an object exceptional if the relative likelihood 
of the value assumed by that object on the attribute is rare if compared to the
relative likelihood associated with the other values assumed
on the same attribute by the other objects of the database.

Let $a$ be an attribute of $A$.
We assume that a random variable $X_a$ is associated
with the attribute $a$, which models the domain of $a$.
Then, with $f_a(x)$ we denote the pdf associated with $X_a$.  The pdf
provides a first indication on the outlierness degree of a given value
$x$, as usually we would expect low pdf values associated to
outliers. However, the sole pdf value is not enoughs.  A given pdf
value represents a hypothetical ``frequency'' for that value in the
sample under consideration. How typical is that ``frequency'' provides
a better insight on the outlierness degree: a low pdf value in a
population exhibiting low values only is not an indicator of an
outlier, whereas an anomalous low pdf value in a population of
significantly higher values denotes that the value under observation
represents an outlier. Thus, analyzing how the values distribute on a
pdf is the key for measuring the degree of outlierness.

Let $X_a^f$ denote the random variable whose pdf represents
the relative likelihood for the pdf $f_a$ to assume a certain value.
The cdf $G_a$ of $X_a^f$ is:
\begin{equation}\label{eq:Gcdf}
G_a(f) = \int_0^f Pr(X_a^f \le f) ~ {\rm d}f. 
\end{equation}

\begin{figure}[t]
\begin{center}
\includegraphics[width=0.495\textwidth]{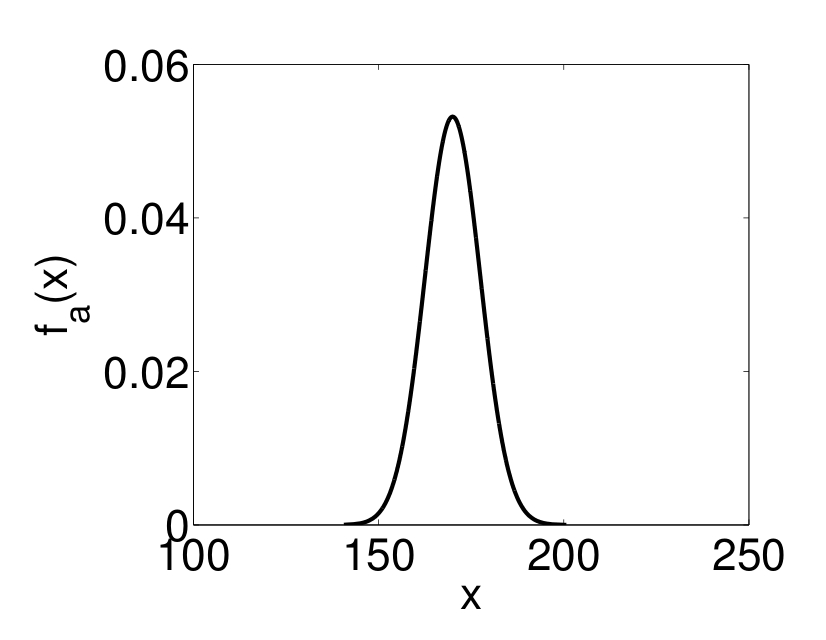}
\includegraphics[width=0.495\textwidth]{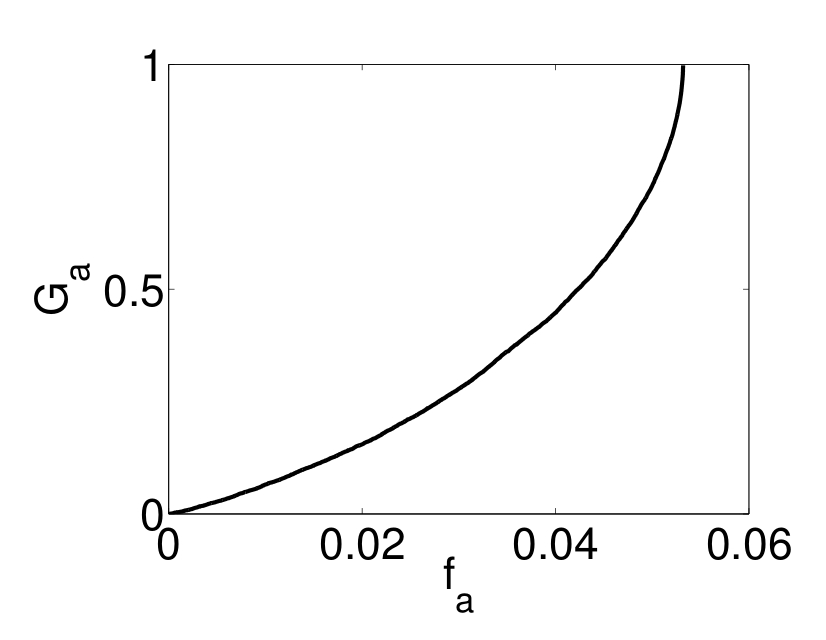}
\end{center}
 \caption{Example of function $G_a(\cdot)$.}
 \label{fig:example_prelim}
 \end{figure}

 \begin{example}
   Assume that the height of the individuals of a population is
   normally distributed with mean $\mu = 170 cm$ and standard
   deviation $\sigma = 7.5cm$. Then, let $a$ be the attribute
   representing the height, $X_a$ is a random variable following the
   same distribution of the domain and $f_a(x)$ is the associated pdf,
   reported in the first graph of fig. \ref{fig:example_prelim}.  The pdf $f_a(x)$ assumes
   value in the domain $[0, f_a(\mu)=0.0532]\subset \mathbb{R}$.
   Consider, now, the random variable $X^f_a$. The cdf $G_a(v)$
   associated with $X^f_a$ denotes the probability for $f_a$ to assume
   value less than or equal to $v$. Then, $G_a(v) = 0$ for each $v\le
   0$ and $G_a(v) = 1$ for each $v\ge 0.0532$.  To compute the value
   of $G_a(v)$ for a generic $v$, the integral reported in Equation
   \eqref{eq:Gcdf} has to be evaluated. The resulting function is
   reported in the second graph of fig. \ref{fig:example_prelim}.
 \end{example}

The \textit{outlierness} ${\rm out}_a(o,\DB)$ (or, simply, ${\rm out}_a(o)$)
of the attribute $a$ in $o$ w.r.t. $\DB$ is defined as follows:
\begin{equation}\label{eq:outlierness}
{\rm out}_a(o) = \Omega \left( \int_{f_a(o[a])}^{+\infty} (1 - G_a(f)) ~ {\rm d}f 
- \int_0^{f_a(o[a])} G_a(f) ~ {\rm d}f 
\right),
\end{equation}

\noindent where $\Omega$ denotes a suitable function mapping 
$\mathbb{R}$ to $[0,1]$
such that ($i$) $\Omega(x)=0$ for $x<0$, and ($ii$) $\Omega$ is 
monotone increasing for $x\ge 0$. In the following we employ the mapping 
\begin{equation*}
\Omega(x) = \frac{1-\exp(-x)}{1+\exp(-x)}.
\end{equation*}
The first integral measures the \textit{area above} the cdf $G_a(f)$ 
for $f> f_a(o[a])$, while the
second integral measures the \textit{area below} the cdf $G_a$ for
$f \leq f_a(o[a])$.
Intuitively, the larger the first term, the larger the degree of unbalanceness
between the occurrence probability of $o[a]$ and that of the values that
are more probable than $o[a]$. As for the second term, the smaller it is,
the more likely the value $o[a]$ to be rare.
Thus, the outlierness value ranges within $[0,1]$ and in particular it
is close to zero for usual properties. By contrast, values closer to one denote
exceptional properties.

\begin{figure}[t]
 \begin{center}
\includegraphics[width=0.495\textwidth]{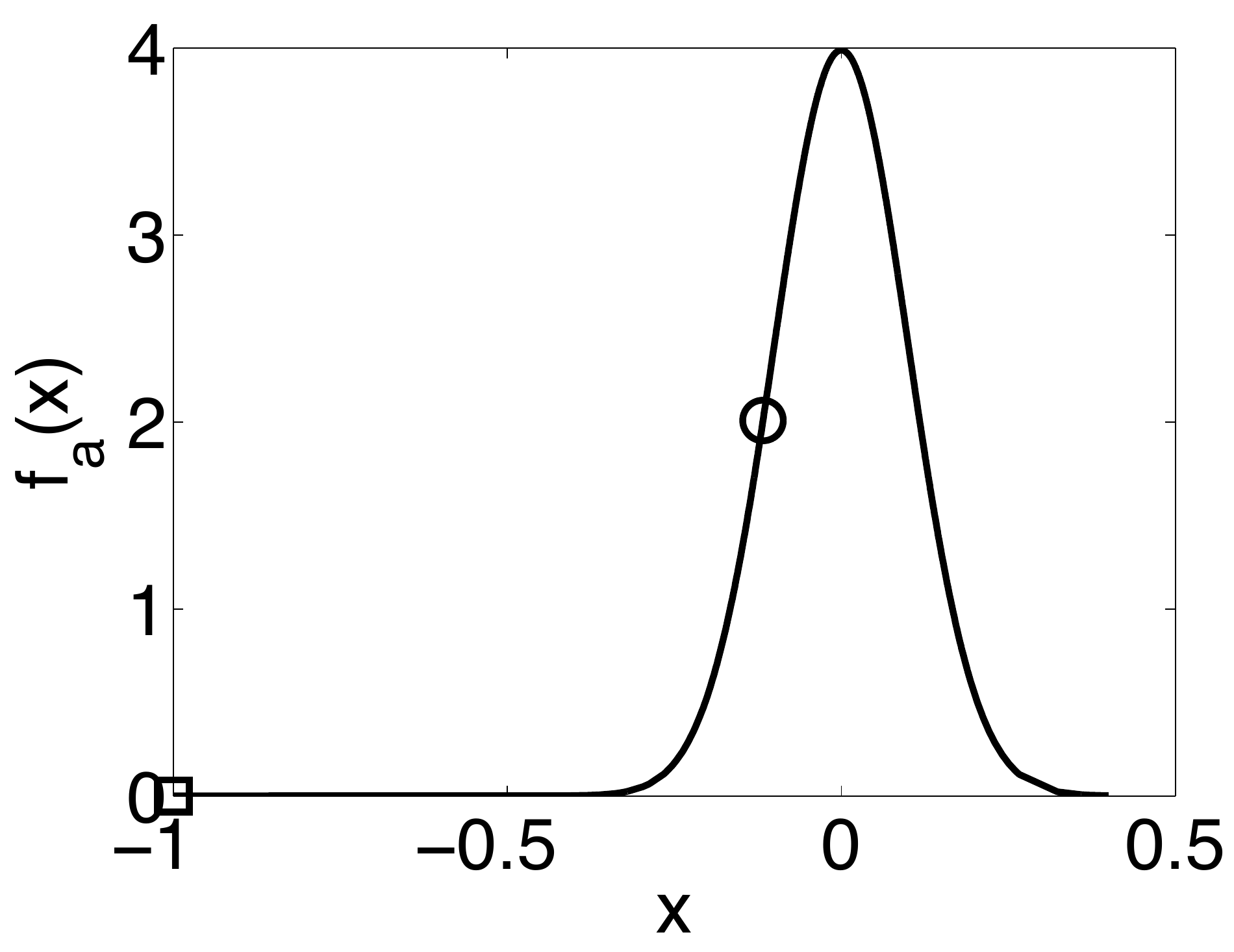}
\includegraphics[width=0.495\textwidth]{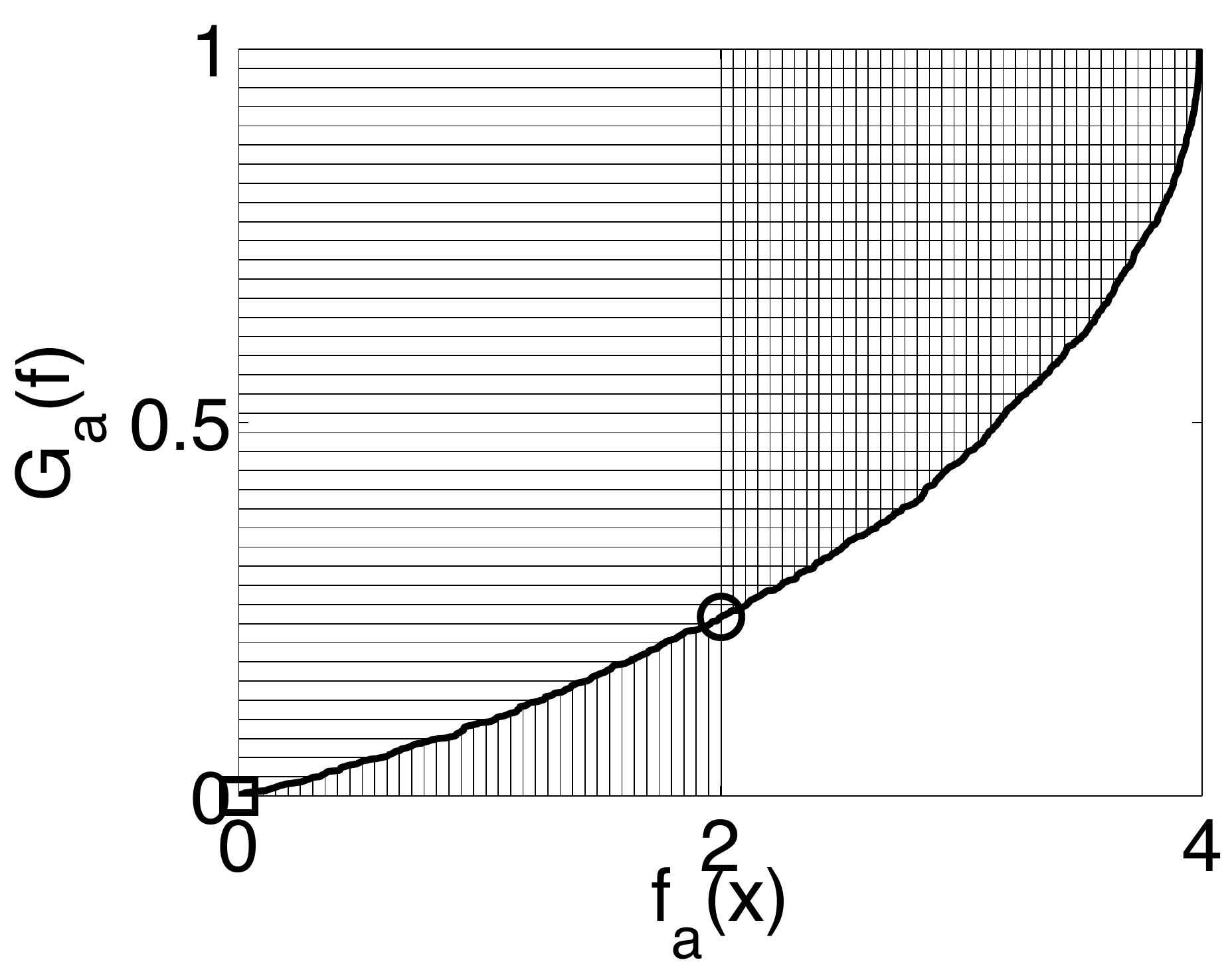}
\end{center}
 \caption{Example of outlierness measure.}
 \label{fig:outlierness}
 \end{figure}

\begin{example}
Consider fig. \ref{fig:outlierness}, reporting  on the left a Gaussian
distribution
$f_a(x)$ 
(with mean $\mu=0$ and standard deviation $\sigma=0.1$).
Consider the values $v_1=-1$ and $v_2=-0.12$, for which
$f_a(v_1)\approx 0$ and $f_a(v_2)\approx 2$ hold.
Assume that an outlier object $o$ exhibits value $v_1$ on $a$.
The associated outlierness $\Out(o)$ corresponds to
the whole area (filled with horizontal lines) above the cdf curve,
that is $\Omega(3.06)=0.91$. 
For an object $o'$ exhibiting value $v_2$ on $a$, instead, the
associated outlierness corresponds to the difference between 
two areas (filled with vertical lines) detected at frequency $2$, 
that is $\Omega(1.17-0.10)=0.49$.
\end{example}

For the sake of clarity, in the above example we considered
a pdf having a simple form. 
However, we wish to point out that our measure
is able to correctly recognize exceptional properties 
irrespectively of the form of the underlying pdf, since it 
compares the occurrence probabilities of the domain values
rather than directly comparing the original domain values.

Given an object $o$ and a dataset $\DB$ on a set of attributes $\mathbf{A}$,
an attribute $p\in\mathbf{A}$ showing a large 
(i.e. exceeding a given threshold)
value ${\rm out}_{p}(o,\DB)$
of outlierness will be called a (\textit{outlying}) \textit{property}
of $o$ in $\DB$.

\subsection{Explanations}
\label{sect:explanations}

\emph{Explanations} are useful in our framework to provide a justification of
the anomalous value characterizing an outlier.
Intuitively, an attribute $a\in\mathbf{A}$ of $o$ that behaves
normally with respect to the database as a whole, 
may be unexpected when
the attention is restricted to a portion of the database. 
Relevant subsets of the database upon which to investigate outlierness
can be hence obtained by selecting the database objects satisfying a
condition, and such that a property is exceptional for $o$.

A condition $c$ (set of conditions $C$, resp.) 
is, intuitively, an \textit{explanation} of the property
$a$, if $o \in \DB_{c}$ ($o \in \DB_C$, resp.) and $a$ is an outlying property
of $o$ in $\DB_c$ ($\DB_C$, resp.). 
Finally, the {\em outlierness} of the attribute $a$ in $o$
w.r.t. $DB$ with {\em explanation} $C$ is defined as
${\rm out}_{a}^C(o,\DB) = \Out(o,\DB_C)$.

It is worth noticing that, according to the relative size of $\DB_C$, 
not all the explanations should be considered equally relevant. 
In the following, we concentrate on $\sigma$-explanations, i.e., conditions
$C$ such that $\frac{|\DB_C|}{\DB} \ge \sigma$,  where
$\sigma\in[0,1]$ is a user-defined parameter.

Thus, given an object $o$ of a database $\DB$ on a set of
attributes $\A$, and parameters $\sigma_\theta\in[0,1]$ and 
$\Omega_\theta\in[0,1]$, the 
problem of interest here is:
\textit{Find the pairs $(E,p)$, with $E\subseteq\A$ and $p\in\A\setminus E$,
such that $E$ is a $\sigma_\theta$-explanation
and ${\rm out}_p^E(o,\DB) \ge \Omega_\theta$.}

The pair $(E,p)$ is also called an (\textit{outlier}) 
\textit{explanation-property pair} of $o$ in $\DB$.

\section{Detecting Outlying Properties}\label{sect:algo}

In order to detect outlying properties and their explanations,
we need to solve two basic problems: (1) computing the outlierness
of a certain multiset of values and (2) determining the conditions to
be employed to form explanations.
The strategies we have designed to solve these two problems 
exploit a common framework, which is based 
on Kernel Density Estimation (KDE).
Specifically, given a numerical attribute $a$, in order to estimate 
the pdf $f_a$
we exploit \textit{generalized kernel density estimation} \cite{JonesH09},
according to which
the estimated density at point $x\in\Dom(a)$ is 
\begin{equation}\label{eq:kde_gen}
\hat{f}_{{\bf m},{\bf w},{\bf b}}(x) 
= \left(\sum_{i=1}^k w_i\right)^{-1} \sum_{i=1}^k \frac{w_i}{b_i} K\left( \frac{x-m_i}{b_i} \right), 
\end{equation}
Here, $K$ is a kernel function, and
${\bf m}=(m_1,\ldots,m_k)$, 
${\bf w}=(w_1,\ldots,w_k)$ 
and ${\bf b}=(b_1,\ldots,b_k)$ are $k$-dimensional vectors
denoting the \textit{kernel location}, \textit{weight}, 
and \textit{bandwidth}, respectively.
The above mentioned strategies are detailed next, together with the
method for mining outlying properties.

\subsection{Outlierness computation}

In order to compute the outlierness,
we specialize formula in Equation \eqref{eq:kde_gen}
by setting ${\bf m} = (x_1,\ldots,x_n)$ and ${\bf w}={\bf 1}$, thus obtaining
\begin{equation}\label{eq:kde}
\hat{f}_a(x) = \frac{1}{n} \sum_{i=1}^n \frac{1}{b_i} K\left( \frac{x-m_i}{b_i} \right), 
\end{equation}
where $x_1, \ldots, x_n$ are the values in $\{ y[a] : y\in\DB \}$,
each term $b_i$ is equal to $h \beta_i$,
with $h$ a global bandwidth (as a rule of thumb,
the global $h$ is set to $1.06\cdot{\rm std}({\bf x})\cdot n^{-1/5}$)
and $\prod_{i=1}^n\beta_i = 1$.

The rationale underlying this choice is that
we want that each value at hand (${\bf m}={\bf x}$) contributes
in equal manner (${\bf w}={\bf 1}$) 
to the estimation of the underlying pdf.
Moreover, we employ the \textit{Parzen window} kernel function,
that is 
$K(x) = 1$, for $|x|\le 1/2$, and $K(x)=0$ otherwise,
since this kernel represents a good trade off between simplicity of computation
and accuracy. 
Indeed, the above density estimate can be computed
in time $O(n\log n)$ by means of a sort of the attribute domain.

We also notice that, since the outlierness depends on the cdf
of the pdf values,
this greatly mitigates the impact of the non-smoothness 
of the estimate of the pdf through Parzen windows,
other than making the measure robust w.r.t. deviations 
of the estimate from the real distribution.

\subsection{Condition building}
\label{sect:algo_cond}

Proper conditions are the  basic 
building blocks for the explanations.
To single them out, our strategy consists in finding, for each attribute $a$, the ``natural'' interval
$I_a$ including $o[a]$,
namely, an interval of homogeneous values on $a$.
Natural intervals, in our modeling, represent a partitioning of
$\Dom(a)$ according to the density $f_a(x)$: intuitively, an interval is a high
density area separated by another interval by a low-density area. 

The search for feasible intervals still relies on adopting the kernel density
family introduced so far, but according to a different interpretation.
In practice, for each attribute $a$, we estimate 
$f_a$ by means of
$\hat{f}_{\mathbf{m},\mathbf{w},\mathbf{b}}$. This latter function can be interpreted
as a mixture density over the parameter sets $\mathbf{m}, \mathbf{w},
\mathbf{b}$. Hence the intervals can be obtained by estimating such
parameters. To this purpose, we adopt the Gaussian kernel 
$
K(x) = \phi(x) =
(2\pi)^{-1/2}\exp(x^2/2)
$ 
and devise the simplifying latent assumption that each data point is generated by a
unique kernel location. This allows us to adopt an EM-based maximum
likelihood approach, 
where the resulting iterative scheme draws from \cite{JonesH09}, and
updates locations and bandwidths according to the following equations: 
\begin{equation}
 m_{j}   =  
\frac{1}{\sum_i \gamma_{ij}} 
 \sum_{i=1}^n
   x_i \gamma_{ij}, \qquad\qquad
 \!\!b_j^2  = 
\frac{1}{\sum_i \gamma_{ij}} 
\sum_{i=1}^n
   \gamma_{ij} (x_i -m_j)^2 \label{eq:centers_amplitudes}
\end{equation}
\noindent Here, $\gamma_{ij}$ represents the mixing probability that value $i$ is
associated with the $j$-th kernel location and, in its turn, is computed at each
iteration as:
\begin{equation}
\gamma_{ij} =
\frac{w_j\phi_{b_j}(x_i-m_j)}{\hat{f}_{\mathbf{m},\mathbf{w},\mathbf{b}}(x_i)} \label{eq:mixing}
\end{equation}
We also adapt the annihilation procedure proposed in \cite{FigueiredoJ02}, 
which allows for an automatic estimation of the optimal number $k^\ast$ of
kernel locations, as well as to ignore the initialization issues. The estimation of the 
parameters is accomplished iteratively for each location$j$, where
each weight is computed as
\begin{equation}
w_j  = \frac{\max \{0,\sum_{i=1}^n \gamma_{ij} - \frac{n}{2}\}}
{\sum_{j=1}^{k^\ast}\max \{0,\sum_{i=1}^n \gamma_{ij} - \frac{n}{2}\}} \label{eq:weights}
\end{equation}
Whenever a weight equals to $0$, the contribution of its component
annihilates in the density estimation. As a consequence, the iterative
procedure can start with a high initial value $k^\ast$, and the
initialization of each mixing probability can be done randomly without
compromising the final result. 
To summarize, the overall scheme can be described as follows: 
\begin{enumerate}
\item Initialize $\gamma_{ij}$ randomly. 
\item For each $j$ compute $w_j$; if $w_j \neq 0$ then update $m_j$
  and $b_j$.
\item Recompute $\gamma_{ij}$ and return to step 2, until the
  improvement in likelihood is negligible. 
\end{enumerate}
The \textit{natural interval} of $o$ in $a$ w.r.t. $\DB$ can be
obtained by exploiting the $\gamma_{ij}$ values. First of all, each
$x_i$ can be assigned to a location $k_i = \arg\max_j
\gamma_{ij}$. Then, let $\overline{k}$ be the location wich $o[a]$ is assigned
to. The interval $I_a$ is then uniquely identified by $[l_a,u_a]$, where $l_a=\min_{i}\left\{x_i \mid k_i = \overline{k}\right\}$ and $u_a=\max_{i}\left\{x_i
\mid k_i = \overline{k}\right\}$.

\hspace{-1em}

\begin{algorithm}[t]
\KwIn{
$o$ : an outlier object\newline 
$\DB$ : a dataset
}
\KwOut{$\cal OP$ : the set of minimal explanation-property pairs of $o$ in $\DB$}
\tcp{First phase}
\ForEach{attribute $a_i\in\A$}{
	Compute interval $I_{a_i}$\;
}
\tcp{Second phase}
\ForEach{attribute $p\in\A$}{
	set $L_1$ to $\{ c_i \equiv a_i\in I_{a_i} \mbox{ s.t. } |\DB_{c_i}|/|\DB|\ge \sigma_\theta \}$\;
	set $j$ to $2$\;
	\While{$j\le k_\theta$ and $L_{j-1}\neq\emptyset$}{
		set $E_j$ to $\{ C\cup\{c\} \mbox{ s.t. } C\in L_{j-1} \mbox{ and } 
		c\in\bigcup L_{j-1} \mbox{ and } c\not\in C \}$\; 
		\ForEach{$C\in E_j$}{
			\If{$|\DB_C|/|DB|\ge \sigma_\theta$}{
				\If{${\rm out}_{p}^C(o,\DB)\ge\Omega_\theta$}{
					set $\cal OP$ to ${\cal OP} \cup\{(C,p)\}$\;
				}\Else{
					set $L_j$ to $L_j\cup\{C\}$\;
				}
			}
		}
		set $j$ to $j+1$\;
	}
}
\Return{$\cal OP$}
\caption{\textit{Outlying Property Detector}($o,a,\DB$)}
\label{algo:mainAlgo}
\end{algorithm}
\hspace{-1em}

\subsection{The mining method}\label{sect:method}

Given a dataset $\DB$ on the set of attributes 
$\A=\{a_1,\ldots,a_m\}$, 
an outlier object $o$,
parameters $\sigma_\theta\in[0,1]$, 
$\Omega_\theta\in[0,1]$, and positive integer $k_\theta\le m$
(representing an upper bound to the size of an acceptable explanation),
the algorithm \textit{Outlying-Property Detector}
computes all the pairs $(E,p)$, with
$|E|\le k_\theta$ and $p \in \A\setminus E$, such that:
\begin{enumerate}
\item
$E$ is a $\sigma_\theta$-explanation, and
\item
the outlierness ${\rm out}_{p}^E(o,\DB)$
is not smaller than $\Omega_\theta$, and
\item
$(E,p)$ is \textit{minimal},
that is there is not a pair $(E',p)$
with $E'\subset E$ for which both points 1 and 2 hold.
\end{enumerate}

The algorithm consists of two main phases.
During the first phase, for each attribute $a_i\in\A$, the interval $I_{a_i}$ and,
hence, the associated condition $a_i\in I_{a_i}$,
is determined by means of the procedure described in Section \ref{sect:algo_cond}.
Given the set of conditions $S = \{a_1\in I_{a_1}, \ldots, a_m\in I_{a_m}\}$ on the $m$ attributes in
$\A$, the second phase 
exploits an apriori-like strategy \cite{AgrawalS94} in order to
search for the pairs $(E,p)$ with 
$E \subseteq S$ meeting the above mentioned conditions.
The computed pairs are accumulated in the
set ${\cal OP}$, which represents the output of the algorithm.

The parameter $k_\theta$ here is introduced in order to 
bind the size of an acceptable explanation. 
As a matter of fact, greater values of $k_\theta$ trigger larger
explanations which are likely to lower the support to unacceptable
values. Also, large explanations result difficult to interpret. 
Notice that by setting $k_\theta$ to the value $m$ all the
pairs can be mined.
In the experimental section we study the effects of the $k_\theta$
parameter on the performances.

As for the cost of the above procedure,
the first step is basically depends on the rate of convergence of the
EM algorithm.  By assuming that the number $k$ of kernel locations is initially
set to $\sqrt{n}$, the basic iteration 
is $O(n^{3/2})$. Notice, however, that 
interval components annihilate early in the first iterations,
so practically we can
assume that the number of intervals $k^\ast$ is bounded to a constant
value. Thus, the overall complexity of the first step is linear in the
size of the data and the number of iterations. 
Clearly, the rate of convergence of the algorithm is also of practical
interest,  and it is usually slower than the quadratic
convergence typically available with Newton-type
methods. \cite{dempster77} shows that the rate of convergence of the EM
algorithm is linear and the it depends on the proportion of
information in the observed data.

As far as the second step is concerned, 
computing the outlierness costs $O(n \log n)$.
Since these two sub-steps are executed at
most $O(m^{k_\theta})$ times, the overall
cost of step 2 is $O(m^{k_\theta} n \log n)$.
However, notice that the apriori-like strategy 
greatly reduces the size of portion of the search space to be explored,
so that the total number of conditions explored in practice is much smaller.

\section{Experimental results}\label{sect:experim}

 We evaluate the technique on both real-life  and synthesized datasets,  with the
 aim of showing the  effectiveness of the proposed  approach. 
The ground truth in such datasets is represented by outlier tuples, 
detected by resorting to the feature bagging algorithm described in
\cite{Lazarevic05}. Briefly, the technique detects outliers by iteratively
running a base outlier detection algorithm on a subset of the
available attributes. Outlier detected in the various runs are then
scored by adopting a \textit{combine} function which assigns a score
to each outlier. 

The bagging technique was instantiated by exploting the base OD
method described in \cite{AngiulliF09}, where the parameters are set
to produce just a single outlier. Further, the \textit{combine}
technique adopted simply scores outliers on the basis of the positive
responses they get within the iterations: if a tuple is
detected as an outlier in a given iteration, it gets a positive
score. Scores are then summarized in the combine function, and tuples
are sorted according to the scores.

The feature bagging technique boosts the robustness of base
outlier detection techniques. By contrast, it is 
difficult to manually infer (e.g., by means of
visualization techniques) justification for outlierness:
A tuple can be reputed an outlier for a
combination of factors which in turn depend on different subsets of
the attributes. 
As a consequence, the analysis of the outliers produced
with such a technique provides a significant benchmark on the
effectiveness of the outlier explanation technique.

We employ three real datasets from the UCI Machine Learning
repository \cite{FrankA10}.
The first two datasets, namely \textit{Ecoli} (with $336$ instances and $7$ attributes) and \textit{Yeast}
(with $1,\!484$ instances and $8$ attributes), contain information about protein localization sites.
The third database, called \textit{Cloud}, contains information about 
cloud cover and includes
$1,\!024$ instances with $10$ attributes. 

The support threshold $\sigma_\theta$ has been set to $0.2$ and the maximum number $k_\theta$
of conditions in the explanation to $3$.
The following table reports the explanation-property pairs
scoring the maximum value of outlierness.

\begin{center}
\begin{tabular}{|@{\,\,}l@{\,\,}|@{\,\,}c@{\,\,}|@{\,\,}c@{\,\,}|@{\,\,}c@{\,\,}|@{\,\,}l@{\,\,}|}
\hline
\multicolumn{1}{|@{\,\,}c@{\,\,}|@{\,\,}}{$\bf DB$} & \bf $\bf o$ & $\bf {out}_p^E(o)$ & $\bf p$ & \multicolumn{1}{@{\,\,}c@{\,\,}|}{$\bf E$}\\
\hline
\hline
\it Ecoli    & 223 & 1.000 & $a_4$ & $\emptyset$ \\
\hline
\it Yeast    & 990 & 0.997 & $a_3$ & $\{ ~ a_2 \in [0.13, 0.38] ~ \}$\\
\hline
\it Cloud    & 354 & 1.000 & $a_6$ & 
$\begin{array}{ll}
\{ & a_1\in [1.0, 6.7], \\ 
& a_2\in [134.9, 255.0], \\
& a_5\in [2,\!450.5,3,\!211.5] ~\}
\end{array}$\\
\hline
\end{tabular}
\end{center}

\begin{figure*}[t]
\begin{center}
 \includegraphics[width=0.495\textwidth]{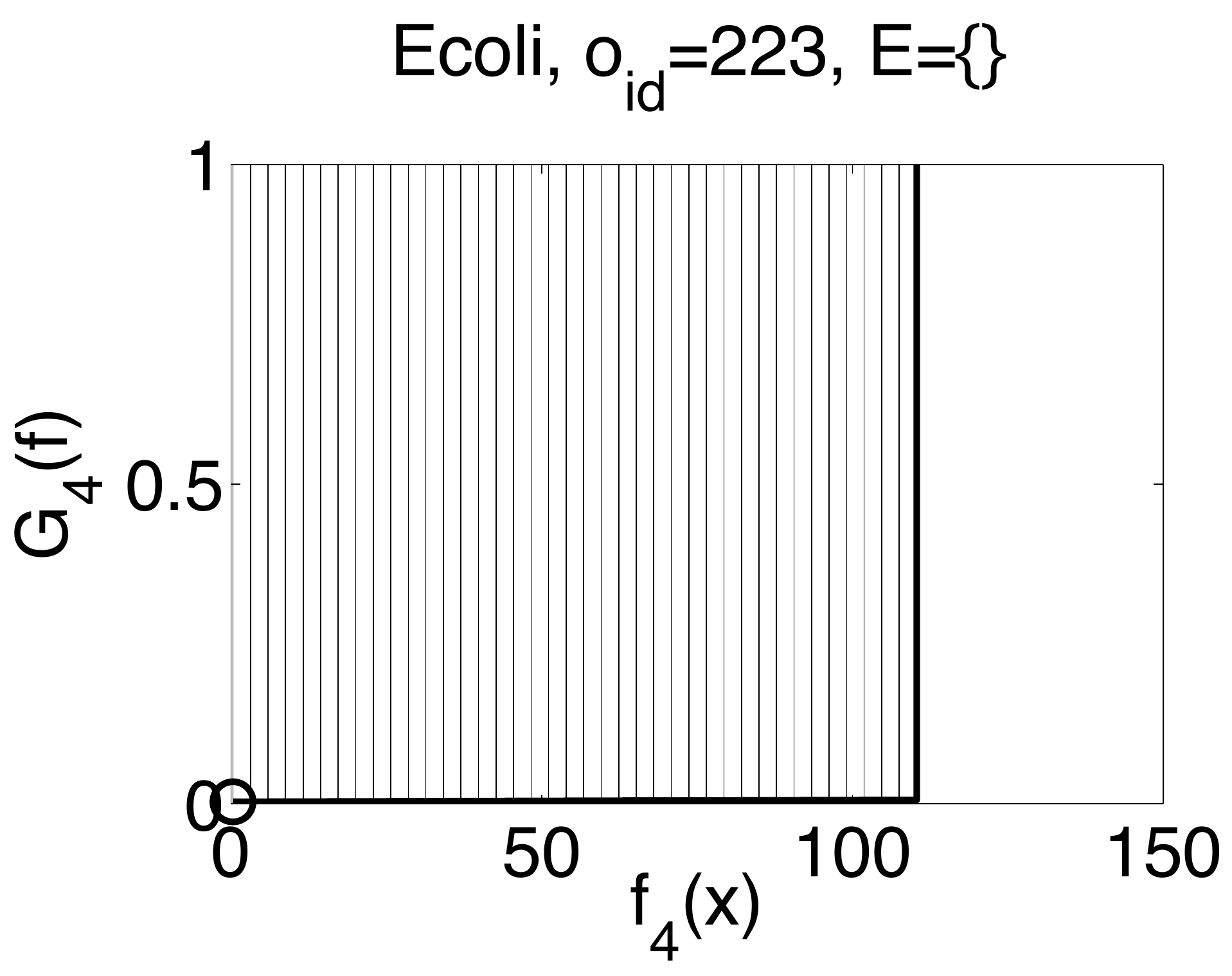}
\includegraphics[width=0.495\textwidth]{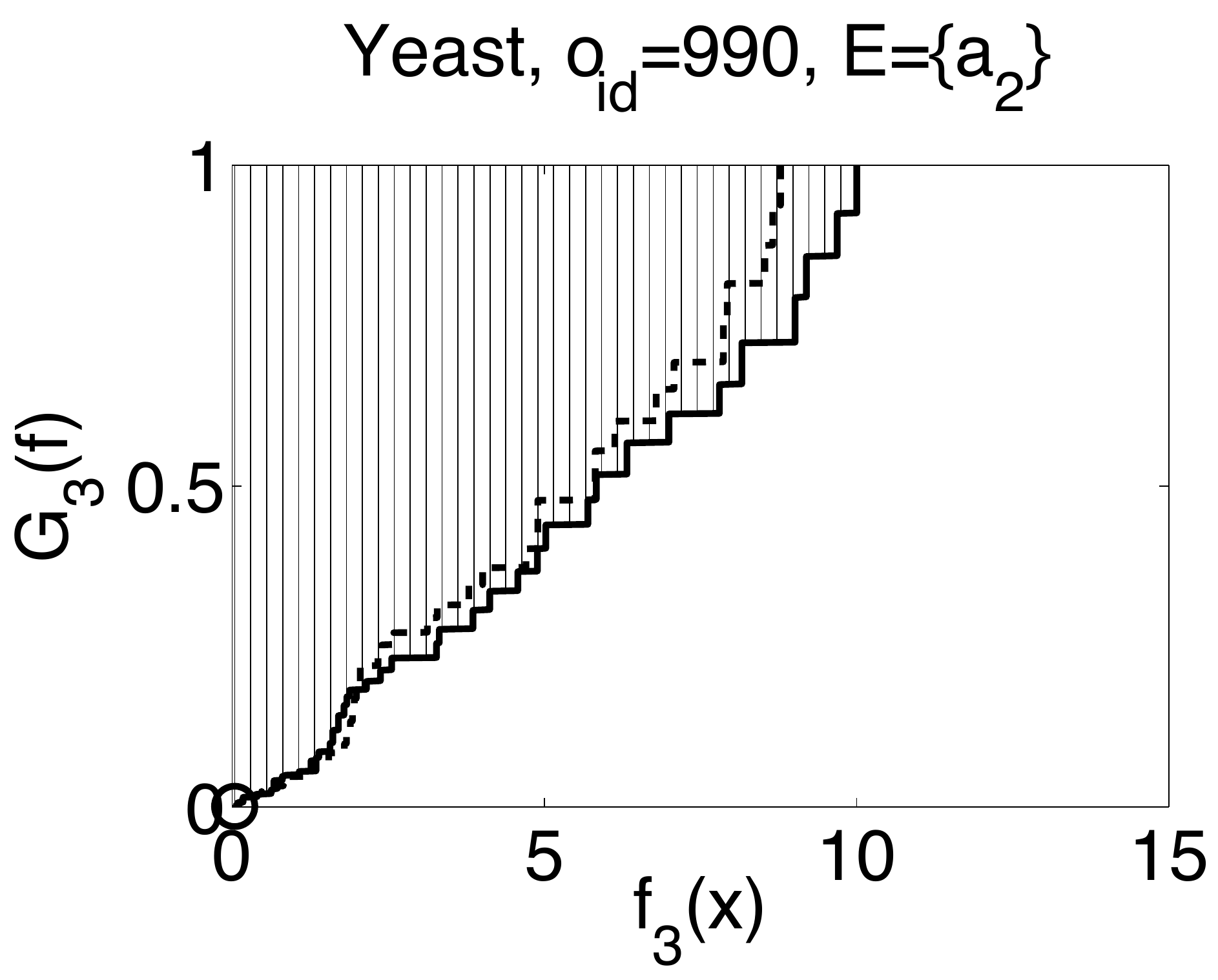} 
 \includegraphics[width=0.495\textwidth]{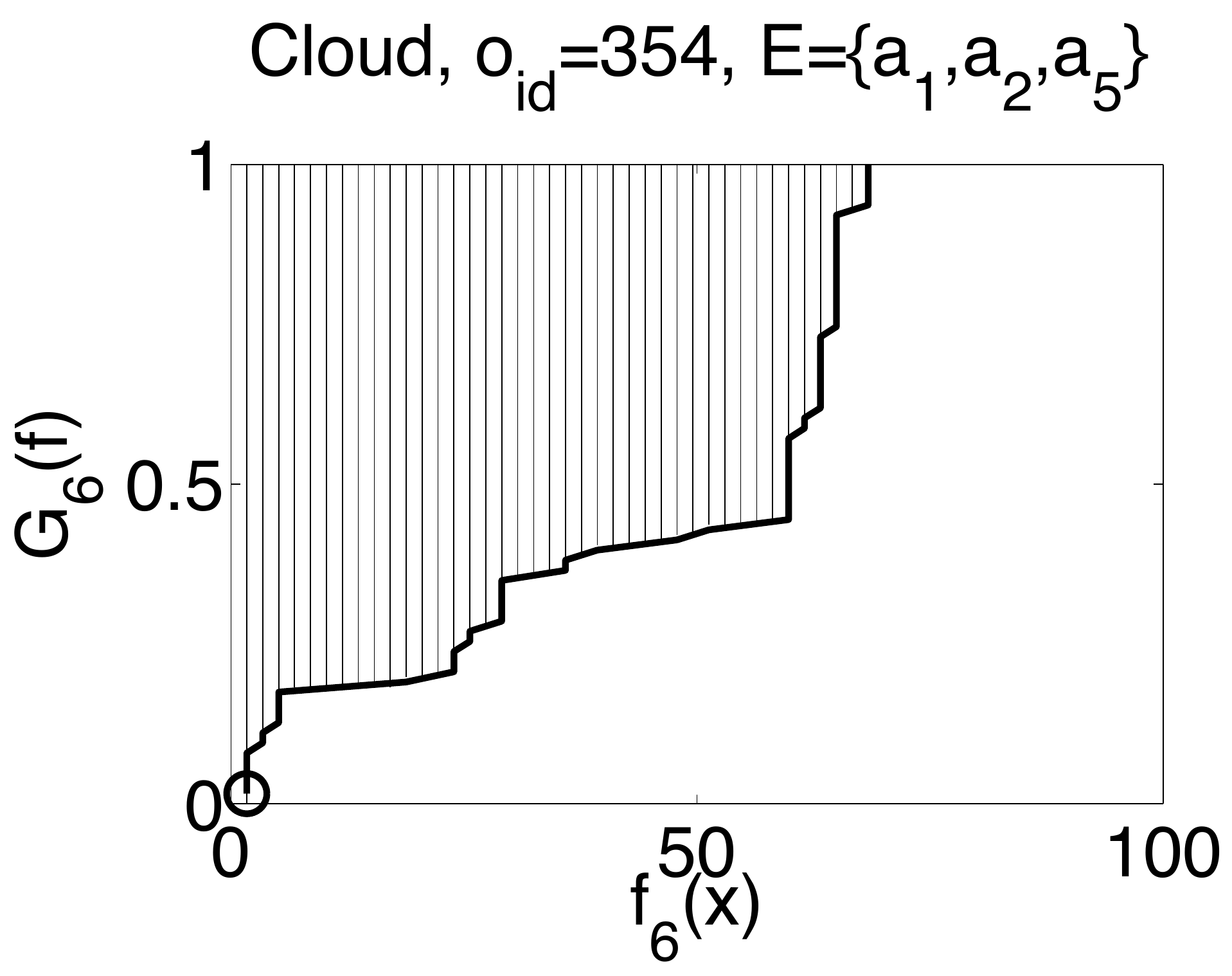}
\includegraphics[width=0.495\textwidth]{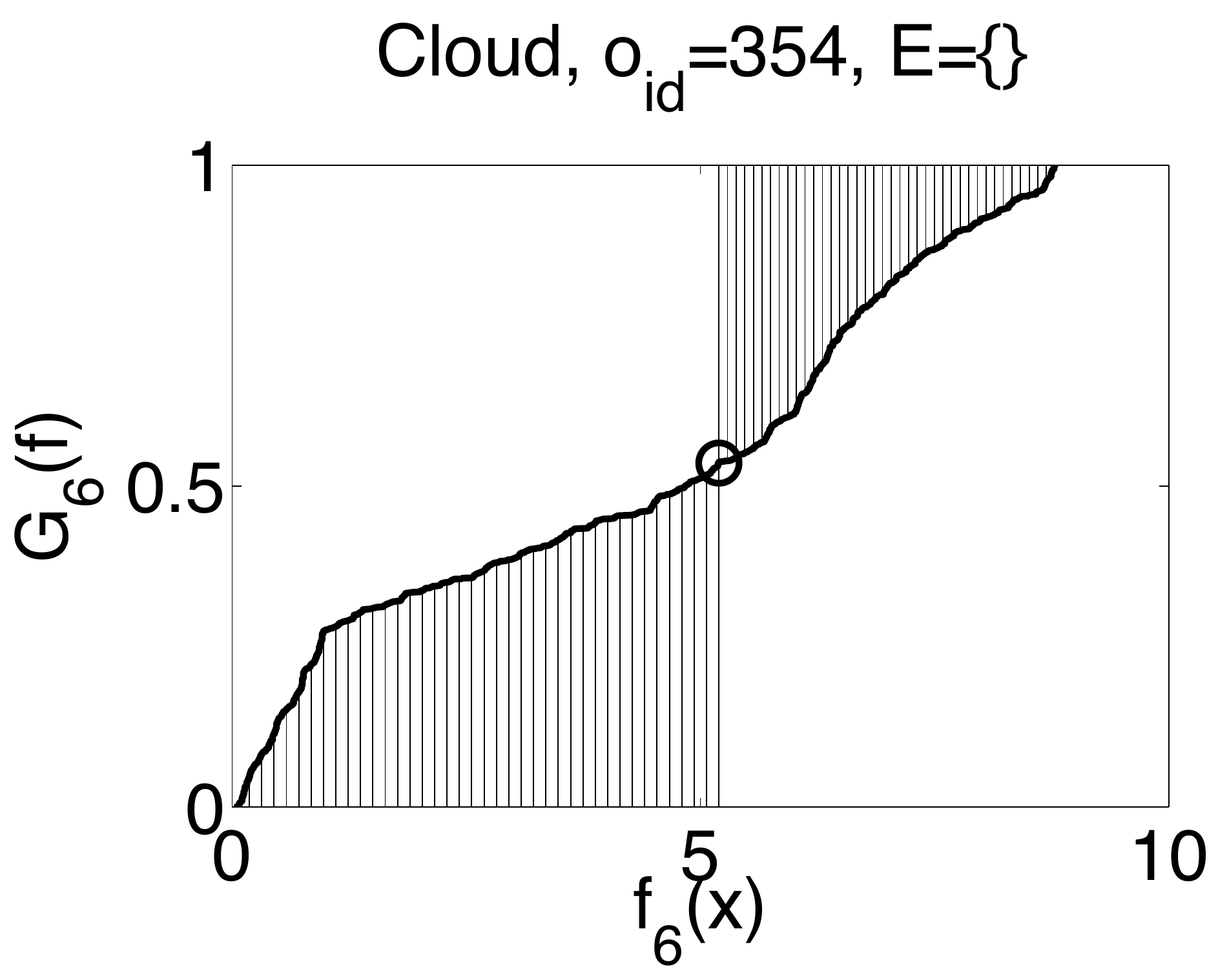}
\end{center}
\caption{Experimental results on the \textit{Ecoli},
\textit{Yeast}, and \textit{Cloud} datasets.}
\label{fig:exp1}
\end{figure*}

In the third column, we report the outlierness value,
in the fourth column the attribute associated with the property, 
and in the fifth column the explanation.
Figure \ref{fig:exp1} reports the functions $G_a(f)$
associated with the objects considered in the experiments.

Figure \ref{fig:exp1} at the top left reports the area associated with the
property $a_4$ and empty explanation for the
object $223$ in the \textit{Ecoli} database. The property $a_4$ is the attribute \textit{Presence of charge on N-terminus of predicted lipoproteins}. 
The object $223$ is the only object assuming value $0.5$ on this attribute, 
while all the other objects assume value
$1.0$. As a consequence, this attribute is a clear outlying property 
with respect to the whole database and, in fact, 
the associated explanation is empty.

Figure \ref{fig:exp1} at the top right reports 
the area associated with the property $a_3$ for the object $990$ in the $Yeast$ database.
The attribute $a_3$ is \textit{Score of the ALOM membrane spanning region prediction program}.
The solid line represents the curve $G_{a_3}(f)$ obtained 
when the explanation relative to attribute $\{a_2\}$ is taken into account, while the dashed line
represents the curve $G_{a_3}(f)$ obtained for the empty explanation.
There is a limited improvement in the significance of the outlierness
degree when the explanation is taken into account, as shown by the
distance in the two lines. 

Things are substantially different with object $354$ in the \textit{Cloud} database. 
Figure \ref{fig:exp1} at the bottom left 
reports the area associated with the property $a_6$ and the explanation $\{a_1, a_2, a_5\}$.
The attribute $a_6$ is the \textit{Visible entropy}, while the explanation attributes are \textit{Visible mean}, \textit{Visible max}
and \textit{Contrast}. 
Figure \ref{fig:exp1} on the bottom right
reports the area associated with the same property, but for an empty explanation.
Clearly, property $a_6$ is not exceptional with respect 
to the whole dataset, but it becomes very exceptional 
with respect to the subpopulation selected by the explanation.

The following table reports the execution times associated 
with the experiments.

\begin{center}
\small
\begin{tabular}{|l|r|r|}
\hline
\multirow{2}{*}{\bf DB} & \multicolumn{1}{c|}{\textbf{Condition}} & \multicolumn{1}{c|}{\textbf{Outlier}} \\
                        & \multicolumn{1}{c|}{\textbf{Building}}  & \multicolumn{1}{c|}{\textbf{Computation}}   \\
\hline
\hline
\it Ecoli       &   6.39 {\it sec} &  16.76 {\it sec} \\
\hline
\it Yeast       &  54.38 {\it sec} & 138.51 {\it sec} \\
\hline
\it Cloud       & 702.67 {\it sec} &  91.08 {\it sec}  \\
\hline
\end{tabular}
\end{center}

\begin{figure}[t]
\centering \includegraphics[width=0.6\textwidth]{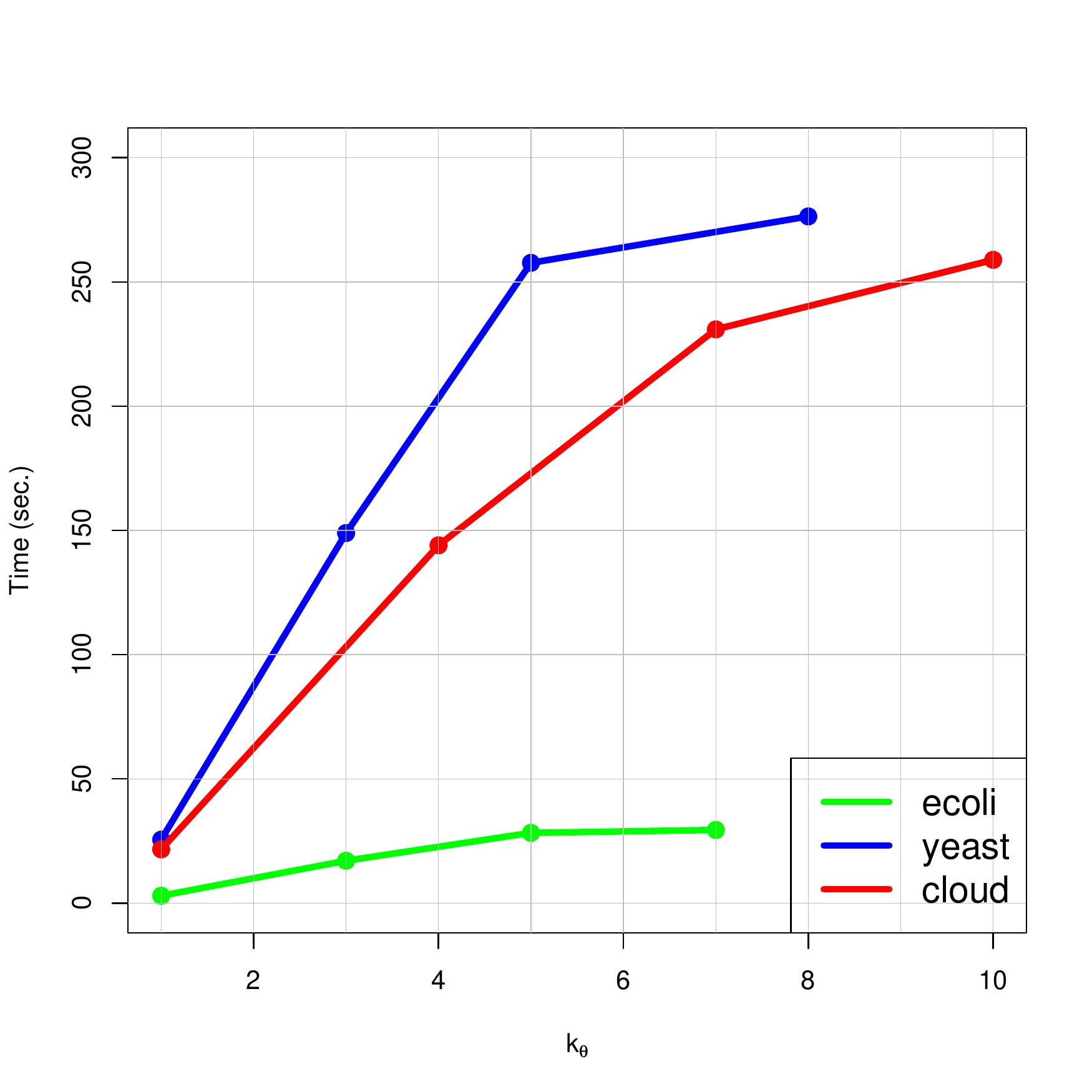}
\caption{Total Outlier Computation Time for \textit{Ecoli},
\textit{Yeast}, and \textit{Cloud} datasets.}
\label{fig:perf}
\end{figure}

\noindent
It can be noticed that the time is split into the two main operations,
namely the identification of intervals, and the computation of the
outlierness degree. The two routines tend to balance the cost of the
overall computation. However, since the parameter $k_\theta$ is likely
to affect the 
performance of the outlier computation, we study the latter on
increasing value of the parameter.  
As a matter of fact, greater values of $k_\theta$ trigger larger
explanations which are likely to lower the support to unacceptable
values.  Figure \ref{fig:perf} plots the
Total Outlier Computation Time
for increasing values of $k_\theta$. The curves tend to flatten for
increasing values, on all datasets, as an affect of the shrinking of
$\mathit{DB}_C$ when $C$ tends to become large.

It is natural to ask whether the computation of the outlierness degree
based on kernel density estimation provides a true advantage over the
alternative approach of first discretizing the attributes, and then
applying the originary method described in \cite{AngiulliFP09}. 
To this aim, we perform further tests on synthesized data. 
In particular, we generate a dataset (named \textit{Unif2} in the
following),  consisting of $20,\!000$ objects.
This dataset contains an outlier $o$
which is distinguished from the rest of the population
from the value it assumes on a particular attribute
$A$.
Specifically, almost all values of this attribute
belong to two equally-sized uniformly distributed clusters,
the first one in the range $[-1.1,-0.1]$
and the second one in the range $[0.1,1.1]$.
The only exception is represented by the object $o$, 
for which $o[A]=0$ holds.
In the following, we concentrate the comparison on the analysis of the
behavior of the two methods on the attribute $A$, 
in order to demonstrate that while $A$ is 
naturally perceived as an outlying property
by the technique hereby introduced, 
it is very unlikely 
to obtain the same goal when the technique
in \cite{AngiulliFP09} is employed.

\begin{figure*}[t]
 \begin{center}
  \includegraphics[width=0.495\textwidth]{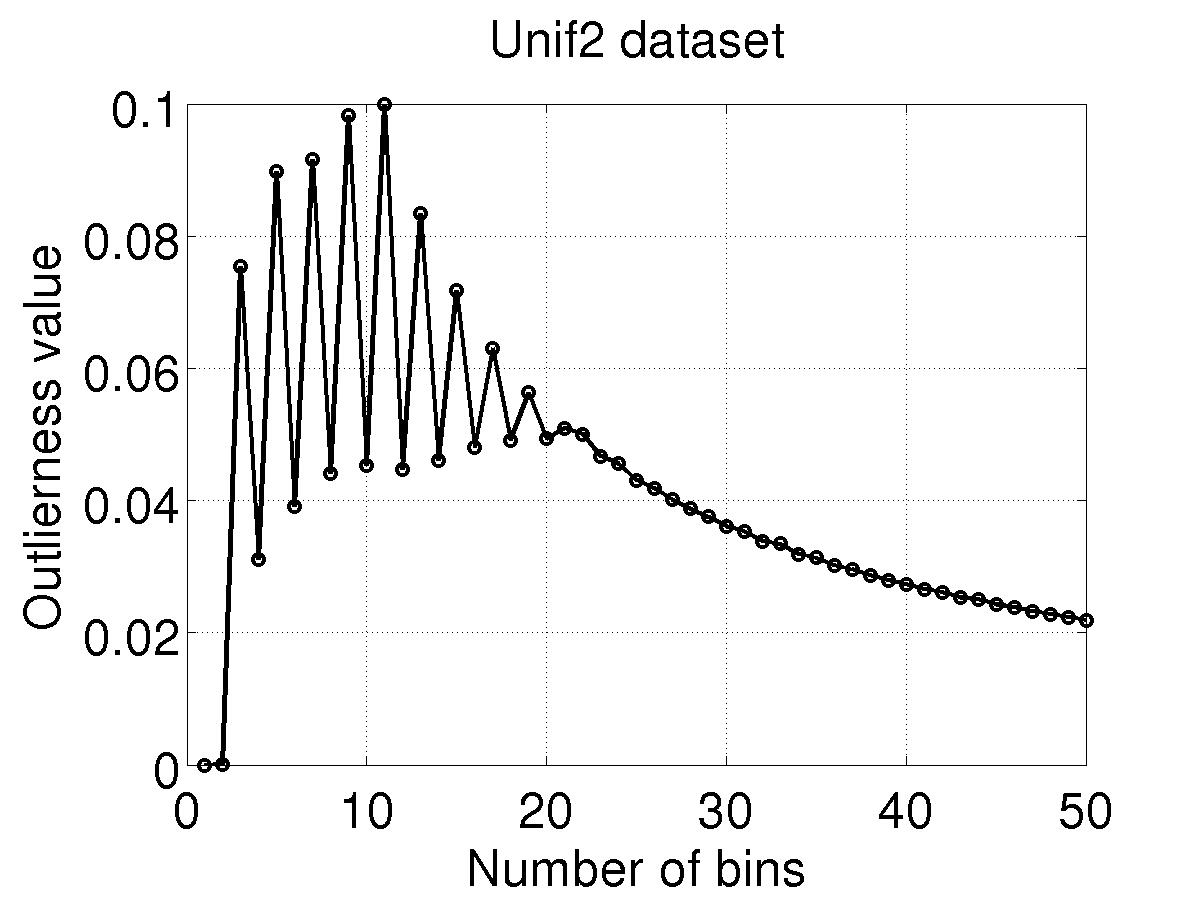}
  \includegraphics[width=0.495\textwidth]{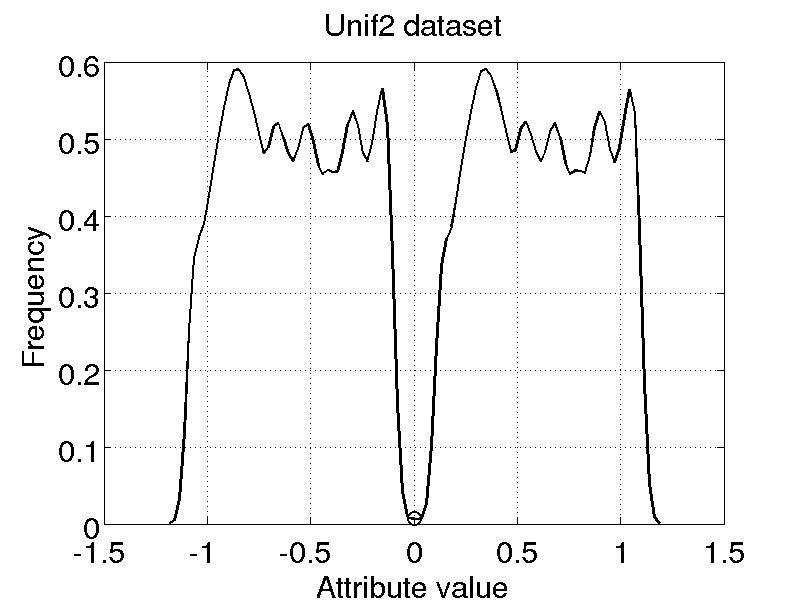}
 \end{center}
 \caption{\textit{Unif2} dataset: outlierness of $A$ in $o$ computed using the method in \cite{AngiulliFP09} (on the left), and density estimate of the same attribute
 carried out by our method (on the right).}
 \label{fig:unif2a}
\end{figure*}

\begin{figure*}[t]
 \begin{center}
  \includegraphics[width=0.325\textwidth]{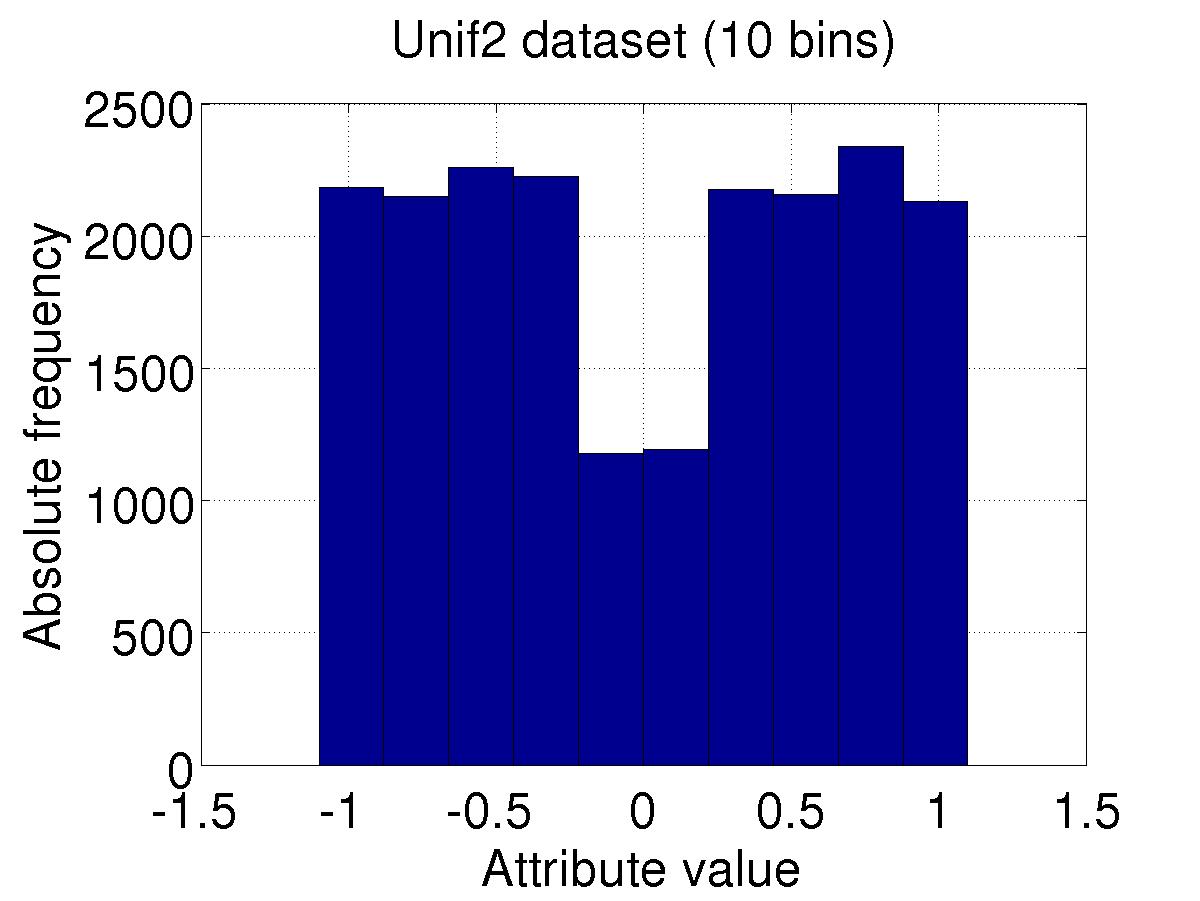}
  \includegraphics[width=0.325\textwidth]{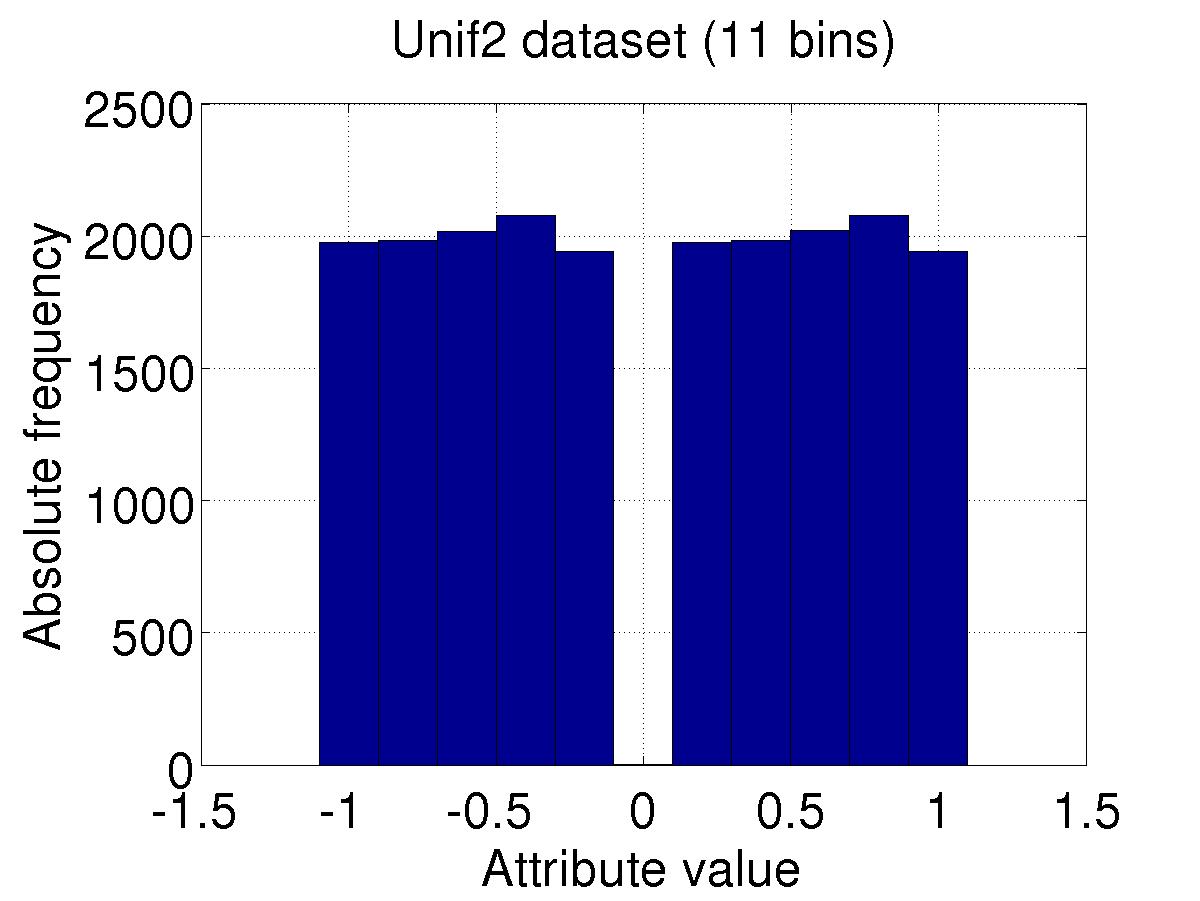}
  \includegraphics[width=0.325\textwidth]{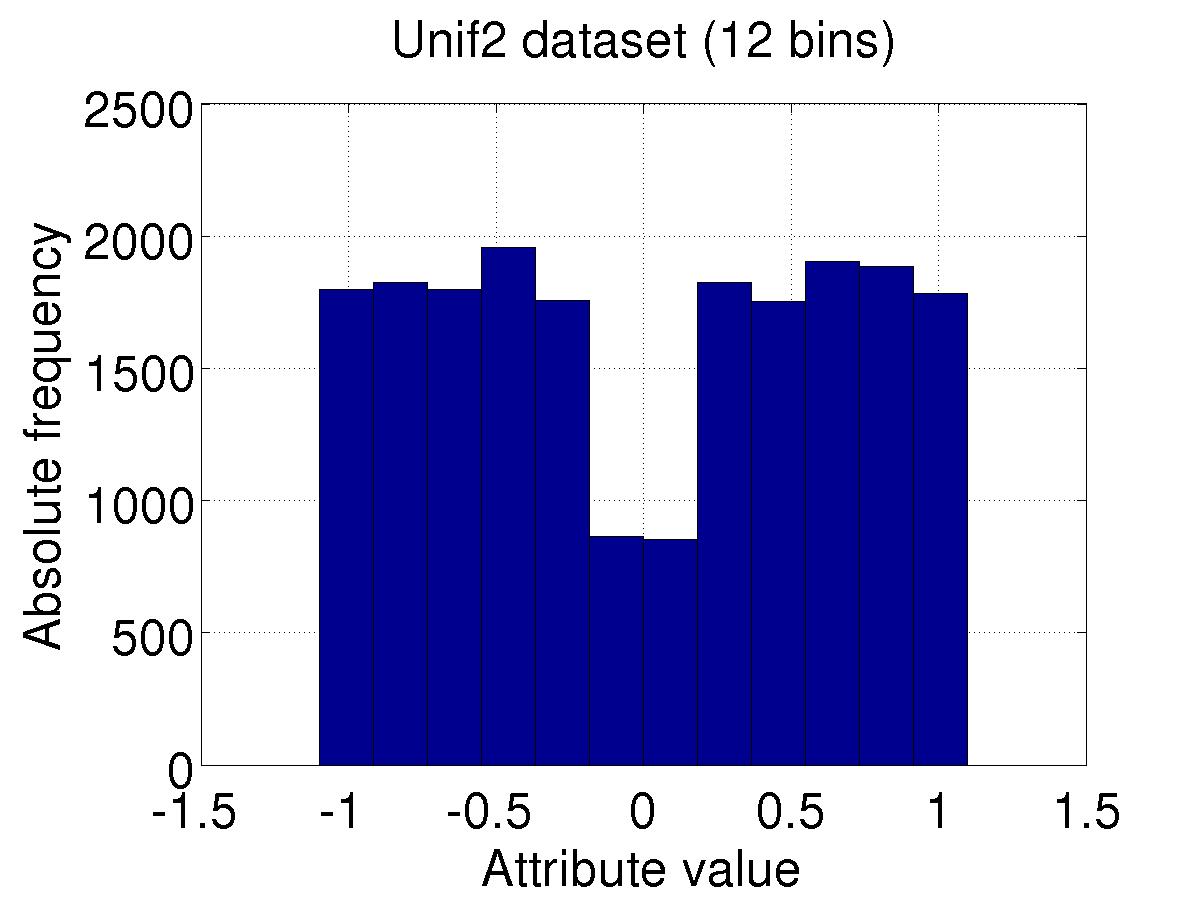}
 \end{center}
 \caption{Different equi-width histograms associated with the attribute
 $A_{U2}$ of the \textit{Unif2} data set.}
 \label{fig:unif2b}
\end{figure*}

In order to apply the latter method
to the \textit{Unif2} dataset, we discretize the attributes 
by grouping attribute values in equi-width bins. 
Figure \ref{fig:unif2a} reports on the left
the value of the outlierness (as defined in \cite{AngiulliFP09})
on the value $o[A]$ 
according to different bins sizes  employed in discretizing
the data. Specifically, the number of bins has been varied
from $2$ to $50$.
The experiment highlights that
when the method in \cite{AngiulliFP09} is applied,
the outcome of the analysis
strongly depends on the discretization adopted.
In particular when the number of bins is in the range $[4,20]$ the
outlierness measure fluctuates between 0.3 and 1. This means
that even small changes in the number of
bins produce results which can dramatically change.
This is a very undesirable property, since determining the right
number of bins for the analysis at hand is a very challenging task.

Figure \ref{fig:unif2b}, showing different frequency histograms 
associated with
the attribute $A$, should further clarify things.
The histogram associated with the best outlierness value, namely
outlierness $0.1$, is the one using $11$ bins 
(at the center of the figure).
In this case, the central bin (centered in zero) scores a low
value of absolute frequency.
Differently, for both $10$ bins (reported on the left in the same figure)
or $12$ bins (reported on the right),
the fact that the  outlierness of $A$ in $o$
is sensibly smaller 
can be explained by looking at the displayed histograms.
In both cases, the value  of $o$ is grouped
with some more frequent values and, hence, the corresponding
outlierness value gets sensibly smaller.

Providing a larger number of bins does not solve the problem: 
as already pointed out,
the scoring functions assigns a score close to $1$ to very unbalanced distributions,
while its value rapidly decreases when frequencies spread.
And, indeed, 
with a large bin size 
the number of different categorical values 
(each associated with a different bin) becomes large, 
and these values score about the same absolute frequency. The
consequence is that the outlierness values get small as well.

We can conclude that 
in order to enable the method \cite{AngiulliFP09} 
to discover meaningful knowledge,   
the bins that maximize the score 
should be detected in the first place.
However, the interaction with explanations (which select subsets of
the overall population) makes it difficult to provide optimal
a-priori intervals, since the distribution of the property attribute
are likely to change when switching from one explanation to another.

This is clearly not the case with the technique proposed in this
paper. Since the outlierness measure defined  here
directly exploits the density estimate of the object value,
it is completely adaptive to numerical data and does not
suffer of the aforementioned drawbacks.
The outlierness computed by our method is $0.775$.
Figure \ref{fig:unif2b} on the right shows the density estimate
of attribute $A$, together with the value associated to $o$
(notice the circle on the curve), which is exploited in order
to compute the outlierness associated with $o$.

\section{Conclusions and Future Work}\label{sect:concl}

The purpose of this paper has been that of devising techniques by
which the outlying properties detection problem can be solved in the
presence of both categorical and numerical attributes, which
represents a step forward with respect to available literature. The
core of our approach has been the definition of a sensible outlierness
measure, representing a refined generalization of that proposed in
\cite{AngiulliFP09}, which is able to quantify the exceptionality of a
given property featured by the given input anomalous object with
respect to a reference data population. Also, we have developed
algorithms to detect properties characterizing the anomalous object
provided in input.  The experimental results we have
obtained confirm that the presented approach is more than promising.

As a matter of fact, there are several application scenarios where the
proposed technique can be profitably applied. 
Further scenarios include rank learning problems like in
\cite{CFGMO10}: there, the problem of 
detecting rules for characterizing individuals who are scored as
exceptional according to a specific scoring function (like, e.g., the
amount of fraud they commit in a fraud detection scenario) is investigated. It is clear that if
exceptional objects are reputed as outliers, then the outlier
explanation technique described in this paper could be exploited as a basic building
block for rule learning in that domain.

As future work, we are interested
in exploring other strategies for generating proper
conditions and in exnteding the experimental campaign.

\bibliographystyle{plain}

\end{document}